# Robot Operating System Compatible Mobile Robots for Education and Research

Alim Kerem Erdoğmuş[1]

kerem.erdogmus@inovasyonmuhendislik.com

Didem Özüpek Taş[1]

didem.ozupektas@inovasyonmuhendislik.com

Mustafa Karaca[1]

mustafa.karaca@inovasyonmuhendislik.com

Dr. Ugur Yayan[1]

ugur.yayan@inovasyonmuhendislik.com

December 23, 2020

## ABSTRACT

The use of mobile robots has inevitably increased in recent years. The increase in the companies that produce products in this field, the popularity of the studies in the robotic field and the technological competence to serve many different areas have revealed this increase in usage. The importance of mobile robots used as health, education, production, logistics, defense industry and space equipment is now more important than before. The fact that robotic education can be reduced to a very young age, the production and coding of simple robots with easily accessible parts is also an important factor in this field. At this point, the effect of educational robots on the spread of robotic technology cannot be denied.

*K*eywords  robotics · mobile robots · robot operating system · autonomous guided vehicles

## 1 Introduction

The role of the ROS middleware in this robotic education and development of robotics technologies is enormous. ROS was developed as an open source meta operating system for robotics software. It can provide many services that can be expected from a standard operating system, including hardware abstraction and software development in a simulation environment without real hardware, control of the software of the devices I call low-level, functionality due to its widespread use, communication transition between operations and management of robotic software packages. It also provides the tools and code libraries needed to obtain, write, and run code on multiple computers.

There are many different robots developed by Inovasyon Muhendislik for the production, development and training of mobile robots. The basic robotics packages of these robots are shared open source and it is possible for anyone who wants to improve themselves in robotics software. These robots are Evarobot, EvaMars, AGV-OTA, AGV-MOTA and ATEKS robots.

The Evarobot has been developed as a robotic education robot. This robot, which can be used in the ROS Noetic interlayer, has a basic sensor infrastructure (sonar, lidar and camera) that may be required for a robot. Its software has been developed entirely using ROS. It has the ability to map and navigate in the mapped environment. It can be used in gazebo environment. Basic exercises for this robot can be reviewed in the next sections. EvaMars is a six-wheel Mars mobile reconnaissance robot developed in the ROS Noetic interlayer for Uplat Virtual Lab, developed by Innovation Engineering. Seven advanced ROS training tasks have been prepared for EvaMars and are planned to be used in advanced robotics training. The feature that distinguishes it from the Evarobot is that it has high mobility and is developed for a possible Mars mission scenario with a single camera and a digger system. Brief information about this scenario can be examined in the next sections. The AGV-OTA robot has been developed as a factory transport robot. It also supports lidar, sonar and camera like Evarobot. It stands out with its transportation feature. The AGV-MOTA robot is the mecanum wheel integrated version of the AGV-OTA. It has 360 degree mobility. Another feature that differs from OTA is that it has multi-depth camera support. ATEKS is a product created within the scope of the smart wheelchair project. It has camera and sonar sensor support. All of the robots mentioned above are open source and ready to use in the simulation environment.

[1] Research and Development Department, Inovasyon Muhendislik Ltd. Sti., Eskisehir, Turkey

# 2 Robots

This section includes promotional contents, simulation images and videos prepared for robots developed and developed by Inovasyon Muhendislik. These robots are as listed below.

## 2.1 EvaMars

EvaMars [1] is a Mars mission training robot developed by Innovation Engineering for UPlat virtual lab. It is modeled to be able to perform exploration missions in the Martian environment, conduct drilling operations and move through the harsh terrain of the environment. A possible Mars exploration scenario was created with EvaMars and robotics training packages were revealed with this scenario.

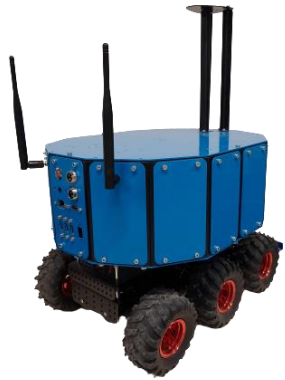

*Figure 1: EvaMars*

EvaMars Robot has a 3-cylinder designed drilling system. The system, which has a piercing tip that can go down to a depth of 15 cm from the surface, is designed to simulate drilling activity scenarios on the Martian surface. Standing out with its mobile aspect, the EvaMars Robot has 6 wheel drive wheels that can move easily in the rough terrain on the Martian surface. For environmental detection, a depth camera with an eye of the Eva-Mars Robot on Mars was used. This camera is integrated for simulations of Mars exploration missions, thanks to both RGB image and depth perception. The robot gains a 180-degree viewpoint, thanks to the rotatable port on which the camera is attached [2].

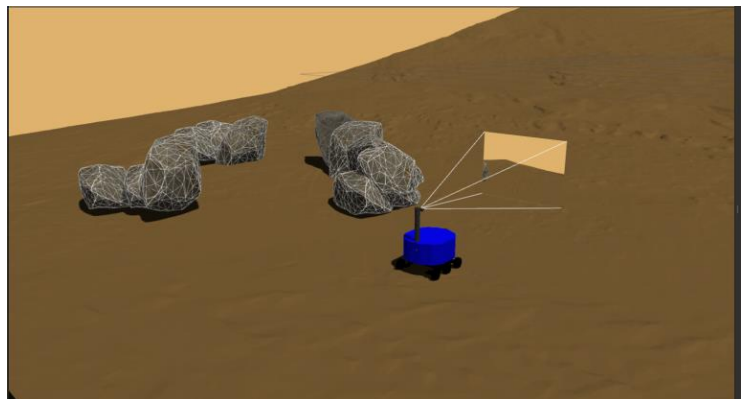

*Figure 2: EvaMars Mars Mission*

## 2.2   Ateks

Ateks [3] is a smart wheelchair developed by İnovasyon Muhendislik.

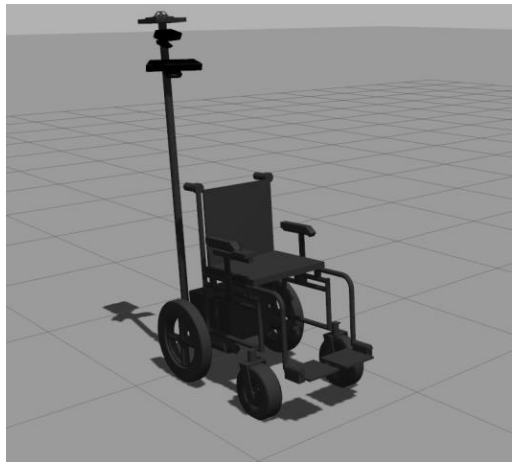

*Figure 3: ATEKS*

Ateks smart wheelchair stands out with its multi-camera support, detecting environmental elements, and having the necessary infrastructure to avoid obstacles with sonar sensor support. It is used for a possible scenario of "patient transport in hospital" in robotics training [4].

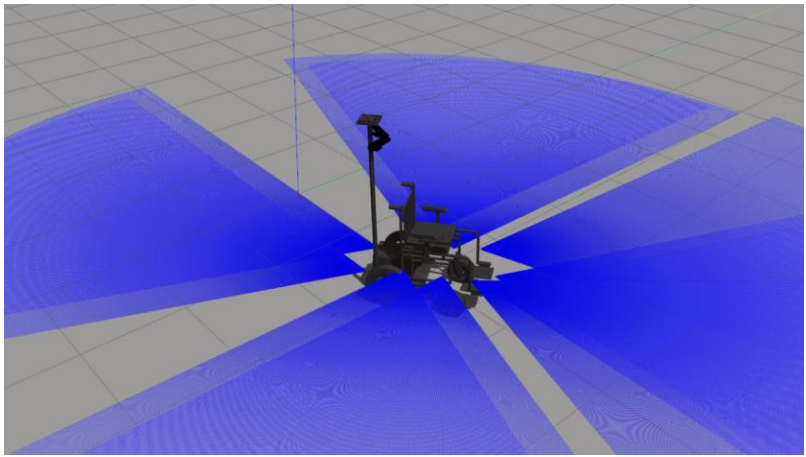

*Figure 4: Ateks Gazebo Environment*

### 2.3 AGV-OTA

AGV-OTA [5] is a factory robot developed by İnovasyon Muhendislik.

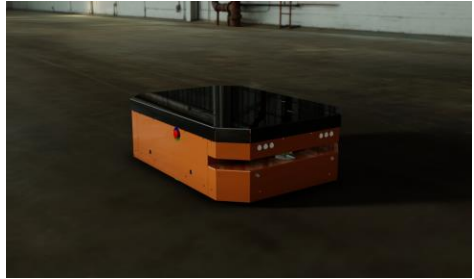

*Figure 5: AGV-OTA*

AGV-OTA stands out with its high carrying capacity and autonomous driving support. It has the ability to walk around the obstacle, route planning, take load by entering under the platform and load carrying capacity up to 100 kg. It is used for a possible "factory robot transport" scenario in robotics training [6].

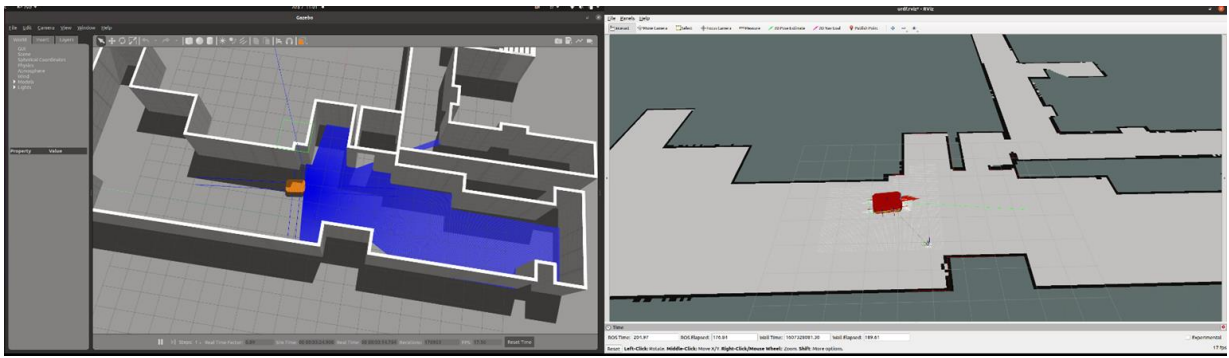

*Figure 6: AGV-OTA SLAM Mapping*

### 2.4 AGV-MOTA

AGV-MOTA [7] is a factory robot developed by İnovasyon Muhendislik.

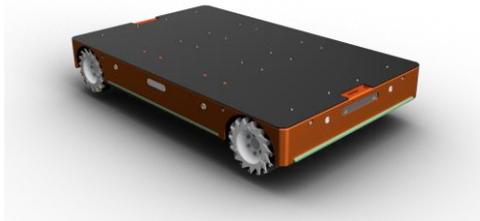

*Figure 7: AGV-MOTA*

AGV-MOTA is the version of AGV-OTA with mecanum wheel integration and increased camera support. Although its transportation capacity has been reduced, AGV-MOTA, whose mobilization and exploration capabilities have been increased, has a 360-degree field of view and 360-degree maneuverability with four separate depth cameras. It also has the feature of recognizing moving and non-moving objects. AGV-MOTA is used for a possible "working office robot" scenario in robotics training [8].

### 2.5 Evarobot/EvaSec

Evarobot [9] is an educational robot developed by İnovasyon Muhendislik [10].

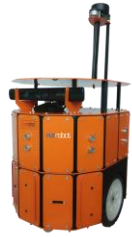

*Figure 8: Evarobot*

EvaSec, on the other hand, is the version of Evarobot designed for basic robotics training, organized in the format of a safety robot. It has been developed for a possible "providing office security" scenario in robotics training. It has camera and sonar support like other robots [11].

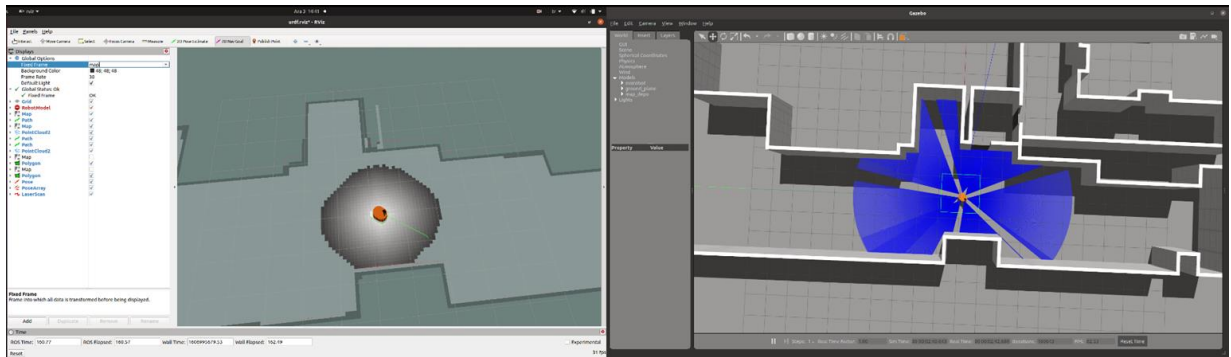

*Figure 9: Evarobot has autonomous driving capability.*

## 3 TUTORIALS

In this section, a sample robotic experiment prepared for robots whose features are briefly mentioned above are explained. These experimental studies have been determined as sample studies that will reveal the distinct features of robots from the experimental sets prepared for robots.

### 3.1 EvaSec Patrol Mission Creation Tutorial

**Mission Purpose:** The node that performs the patrol function of EvaSec should be created and run. The following figure shows the patrol node block diagram of the robot.

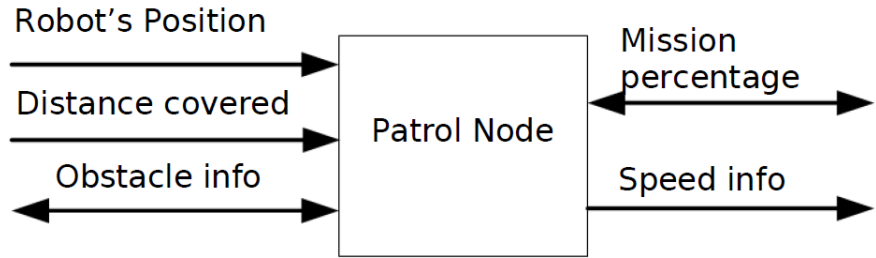

*Figure 10: Patrol Mission Node Diagram*

**Task Expectations:**

- This study is one of the last tasks of the EvaSec test suite. Until this task, the user who has written the movement nodes, obstacle avoidance and publisher-subscriber nodes of the EvaSec robot is now expected to create a patrol task software that will allow the robot to use these nodes in a coordinated manner. The main purpose of the EvaSec security robot's experiment package is to examine the basics of the patrol feature, which is the basic task of the robot, and to activate it.

- While creating this software, the user who is expected to use the Python code language should add the patrol mission parameters (regulating the basic parameters such as head angle and angular velocity) to the "params" folder for the basic parameters supporting this code file, and finally, the points where the robot must go during the patrol it is expected to create the task points parameter file.

- Experiment work is completed when the user writes a launch file that will run this code with parameter files.

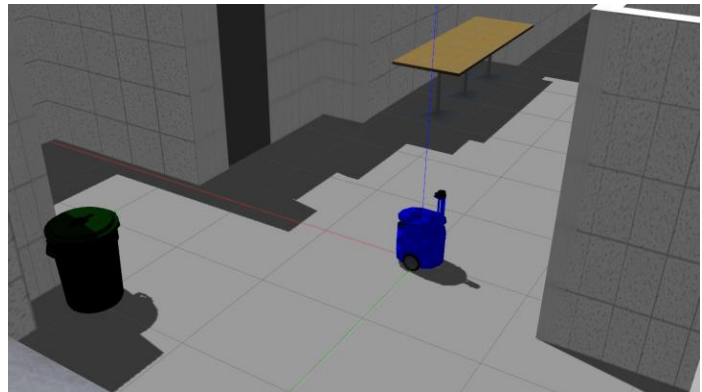

*Figure 11: EvaSec Patrol Mission*

### 3.2    EvaMars Drill Mission Tutorial

**Mission Purpose:** In this mission, which is the last of the Mars missions, it is aimed to use the drilling system that was activated in the previous missions. A scenario in which certain drilling points were discovered with the long expedition route and investigations carried out previously. The robot needs to go to two of the previously discovered spots, drill and return to the point where it was landed on Mars. The user is expected to create a code file to complete this task.

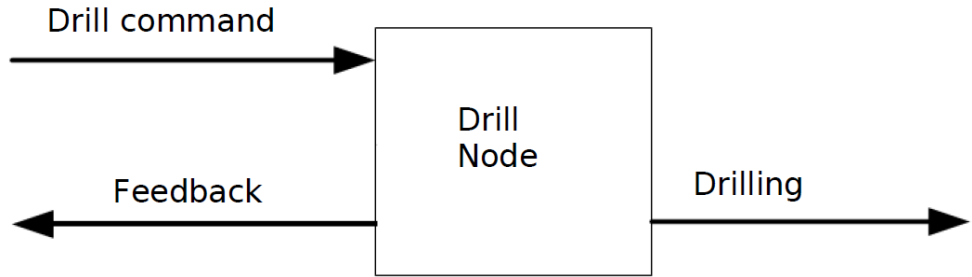

Figure 12: Drill Mission Node Diagram

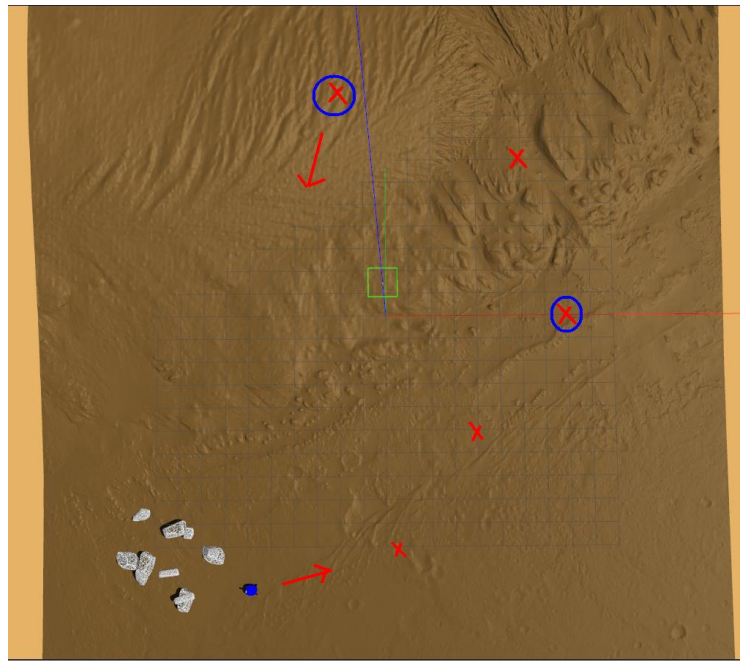

Figure 13: EvaMars Drill Locations

**Task Expectations:**

• This mission pack is designed to make the EvaMars robot, which will perform a possible exploration mission scenario on the Mars surface, to full capacity by the user and complete these tasks. The task package includes many robotic works such as manual/autonomous movement, camera controls, activating the drilling attachment, creating packages that enable the camera depth feature to be converted to laser scanning. In this report, the drilling task example is discussed. The user is asked to use the drilling device, which is already on the robot and activated in previous tasks, at the target points given to the user.

• The robot must go to these points autonomously, drill at the points it reaches, go to the next point when the drilling process is completed and repeat the same process.

• The task is completed with the creation of the Python code file that will perform this work, the completion of the drilling process and the necessary reporting.

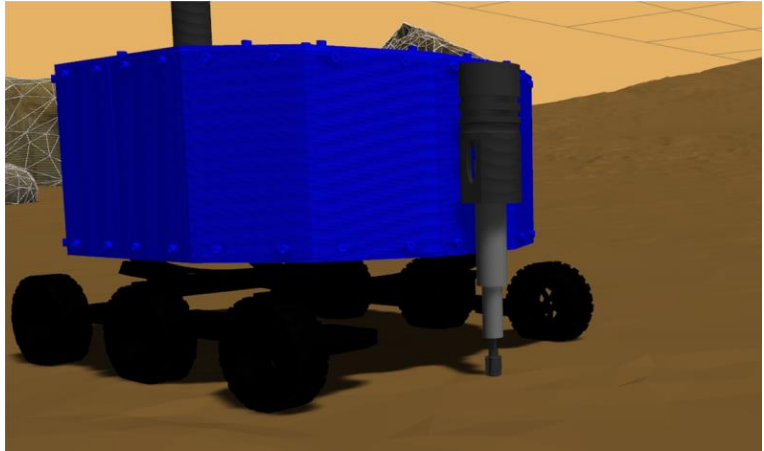

*Figure 14: EvaMars Drill Tool*

### 3.3 AGV-OTA Navigation Tutorial

**Mission Purpose:** It is aimed to carry out the task of transferring the AGV-OTA to the points determined via Rviz in the sample work area mapped in previous missions. For this basic task, the user must install the necessary ROS navigation packages and create the relevant subpackages for the ROS packages of the AGV-OTA.

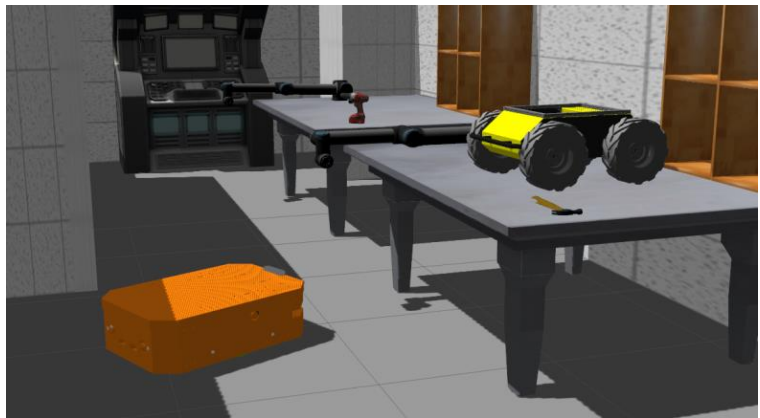

*Figure 15: AGV-OTA Workhouse*

**Task Expectations:**

• This task belongs to the "navigation" part of the standard ROS training package prepared for AGV-OTA. In these training tasks, there are tasks such as calling the standard robot to Gazebo, using Rviz, using teleop and SLAM mapping. In the navigation part of this task tree, the user is expected to use the navigation packages and to perform activities such as creating the necessary subpackage and creating a launch file.

• The task is completed when the "agv_navigasyon" subpackage is operated properly and when it is transferred to different points using the "2D Nav Goal" function of Rviz in the mapped environment.

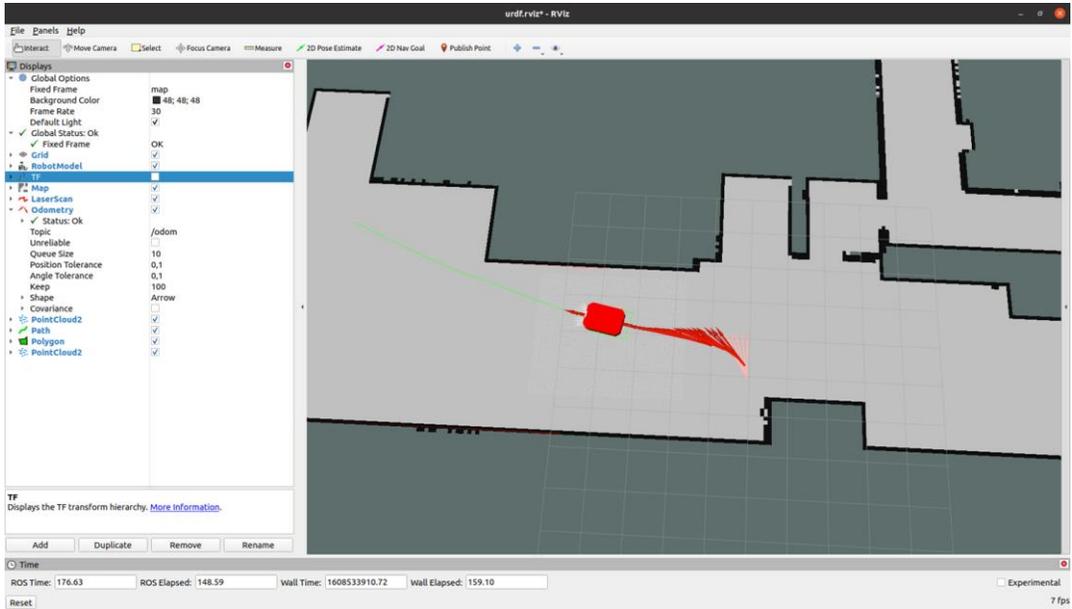

*Figure 16: AGV-OTA RViz Navigation*

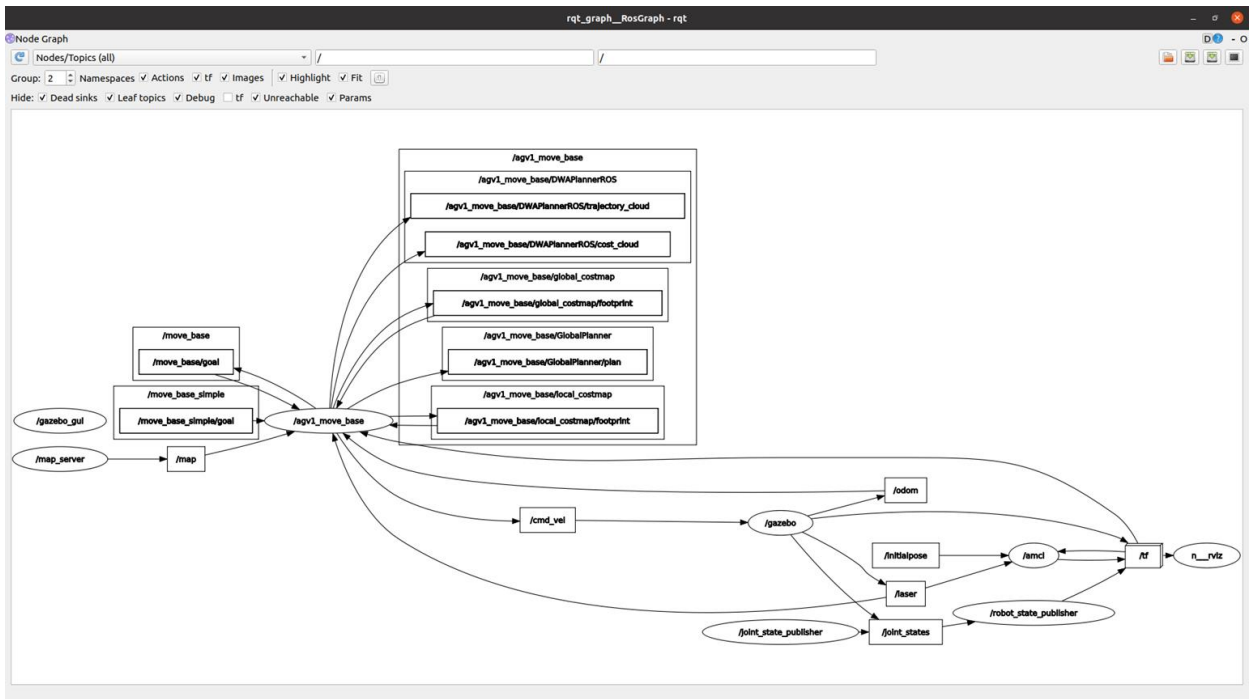

*Figure 17: rqt_graph of AGV-OTA*

### 3.4  AGV-MOTA Environment Exploration Mission Tutorial

**Mission Purpose:** Using the multi-camera support of the AGV-MOTA, it is aimed to view a working environment, capture images, create point cloud visuals with depth cameras and examine them through Rviz.

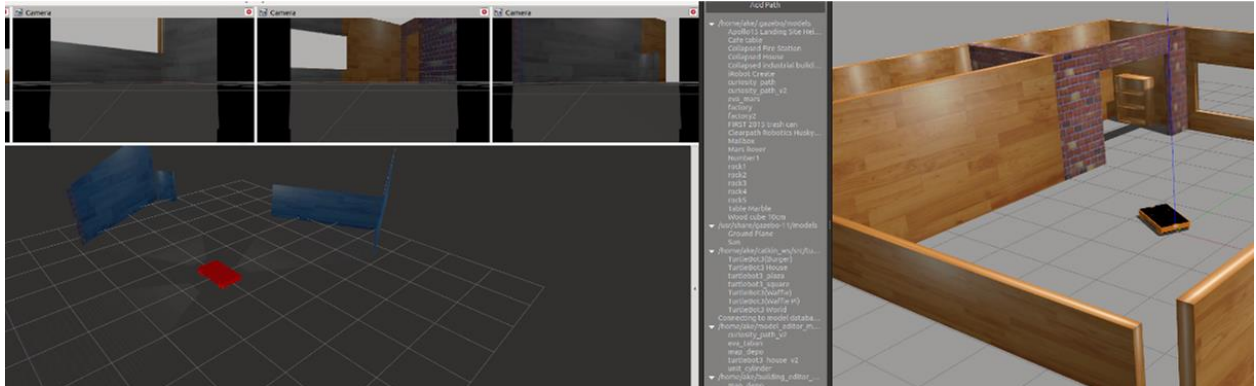
*Figure 18: AGV-MOTA Environment Exploration*

**Task Expectations:**

• This task belongs to the "camera and imaging" part of the standard ROS training package prepared for AGV-MOTA. These training tasks include calling the standard robot to Gazebo, using Rviz, using teleop and using depth camera of AGV-MOTA. In the camera and imaging part of this task tree, the user is expected to perform activities such as using the robot's cameras for observation over Rviz, recording images with the image_saver package, and using the PointCloud tool of Rviz.

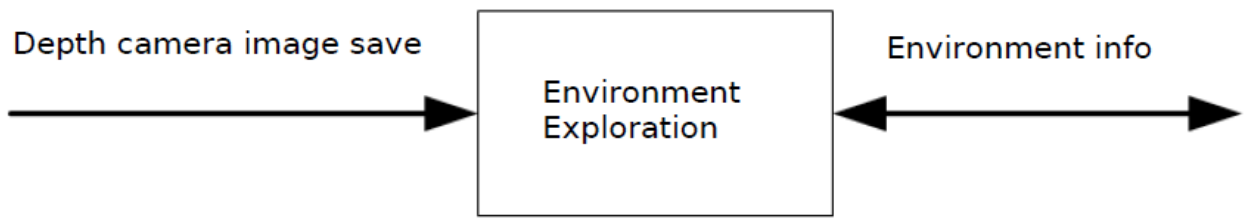
*Figure 19: Environment Exploration Diagram*

### 3.5 ATEKS Obstacle Avoidance Mission Tutorial

**Mission Purpose:** It is aimed to develop a Python code that processes the obstacle distance data of the sonar sensors owned by ATEKS and provides progress without hitting obstacles.

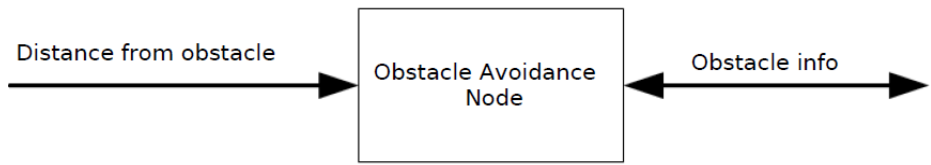
*Figure 20: Obstacle Avoidance Node Diagram*

**Task Expectations:**

• The short training package developed for ATEKS is based on the scenario of effective use of a smart wheelchair in a hospital environment. The use of the obstacle avoidance algorithm, which is the last task of this package, aims to create a task content that simulates the patient using ATEKS to return to his room without any problems. The user is expected to be aware of the obstacles of the chair by processing the sensor data of ATEKS and to reach the hospital room without hitting the corridor.

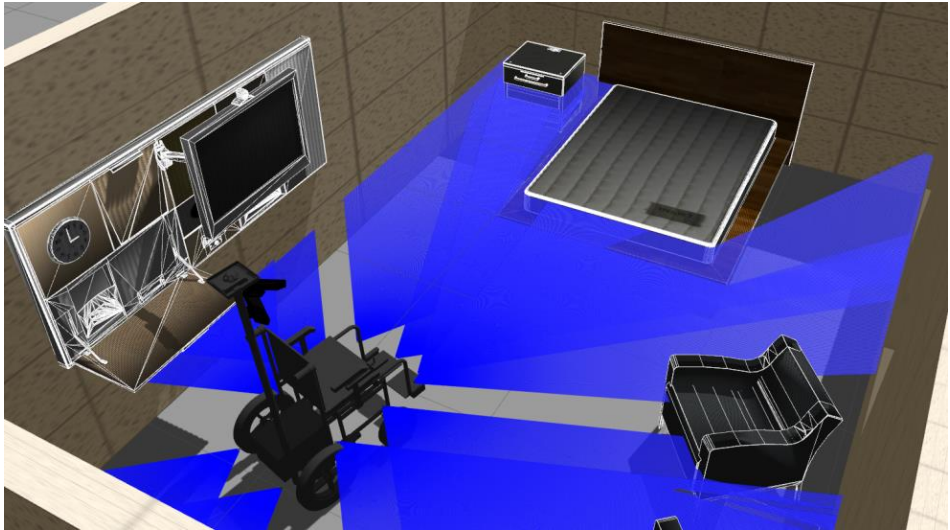

*Figure 21: ATEKS Obstacle Avoidance Mission in Hospital Environment*

### 3.6 Learning Outcomes

As can be seen from the sections of the robotic training scenarios given in the previous section, the training scenario designed for each robot is about the user having knowledge about the ROS interface and the use of robots designed in this interface, developing and using the code in Python language on this system. The achievements of these training scenarios are listed below.

**EvaSec – Patrol Mission Tutorials:**

• This task series has been developed for first-time ROS users with beginner knowledge of ROS and Python.

• At the end of this series of tasks, users acquire beginner-level ROS knowledge such as creating and using basic ROS packages, creating basic message / service packages, writing simple scripts, viewing and using a robot in a simulation environment.

**EvaMars – Mars Mission Tutorials:**

• This task series has been developed for users with intermediate knowledge of ROS and Python who have previously used any robot on ROS.

• At the end of this series of tasks, users will have developed intermediate level ROS knowledge such as developing complex nodes on ROS, writing advanced Python code files and using them on ROS, creating codes and nodes that will effectively use the sensors owned by a robot, and developing autonomous driving algorithms.

**AGV-OTA – Factory Mission Tutorials:**

• This task series has been developed for users with intermediate knowledge of ROS and Python who have previously used any robot on ROS.

• At the end of this task series, users will acquire the use of a factory robot that can be used in a real factory environment thanks to ROS and Gazebo support, node and package developments for this robot, basic navigation and SLAM mapping knowledge, intermediate ROS knowledge such as autonomous driving.

**AGV-MOTA – Workhouse Exploration/Work Tutorials:**

• This task series has been developed for users with intermediate knowledge of ROS and Python who have previously used any robot on ROS.

• At the end of this task series, users will be able to use a robot that can be used in a real workshop environment thanks to ROS and Gazebo support, node and package development for this robot, the use of multi-camera support, medium-level ROS such as enabling 3D viewing and recording of the environment thanks to the depth camera. will have the knowledge.

**ATEKS – Hospital Missions Tutorial:**

• This task series has been developed for first-time ROS users with beginner knowledge of ROS and Python.

• At the end of this series of tasks, users acquire beginner ROS knowledge such as creating and using basic

ROS packages, creating basic message / service packages, writing simple scripts, using a smart wheelchair that can be used in a real hospital environment thanks to ROS and Gazebo.

In line with the gains above, the order of ROS training scenarios from simple to difficult is shown in the diagram below.

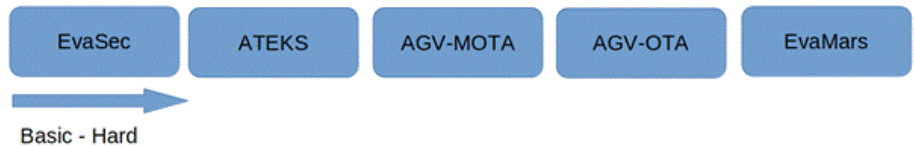

*Figure 22: Mission Difficulties*

## 4  Conclusion

This study is a resource that includes the introduction of all mobile robots developed by Inovasyon Muhendislik and one example from the training packages created with the simulation models of these robots. These experimental studies are designed to serve the necessity of increasing the number of people trained in this field, with the prediction that the use of mobile robots in human life will increase in the future. All of the experiment packages and all robot packages that are not included here will be uploaded to the Uplat virtual laboratory environment [12] and will be made accessible and used by anyone who wants to work and improve themselves in this field.